\documentclass{article}

\usepackage{PRIMEarxiv}

\usepackage[utf8]{inputenc} 
\usepackage[T1]{fontenc}    
\usepackage{hyperref}       
\usepackage{url}            
\usepackage{booktabs}       
\usepackage{amsfonts}       
\usepackage{nicefrac}       
\usepackage{microtype}      
\usepackage{lipsum}
\usepackage{fancyhdr}       
\usepackage{graphicx}       
\graphicspath{{media/}}     

\title{Res-CNN-BiLSTM Network for overcoming Mental Health Disturbances caused due to Cyberbullying through Social Media}

\author{
  Raunak Joshi \\
  Research Scholar \\
  University of Mumbai \\
  Mumbai, India\\
  \texttt{raunakjoshi.m@gmail.com} \\
   \And
  Abhishek Gupta \\
  Department of EXTC \\
  University of Mumbai \\
  Mumbai, India\\
  \texttt{abhishekgupta20001@gmail.com} \\
   \And
   Nandan Kanvinde \\
   Depatment of MCA \\
   University of Mumbai \\
   Mumbai, India \\
   \texttt{kanvindenandan81@gmail.com} \\
}

\begin{document}
\maketitle

\begin{abstract}
Mental Health Disturbance has many reasons and cyberbullying is one of the major causes that does exploitation using social media as an instrument. The cyberbullying is done on the basis of Religion, Ethnicity, Age and Gender which is a sensitive psychological issue. This can be addressed using Natural Language Processing with Deep Learning, since social media is the medium and it generates massive form of data in textual form. Such data can be leveraged to find the semantics and derive what type of cyberbullying is done and who are the people involved for early measures. Since deriving semantics is essential we proposed a Hybrid Deep Learning Model named 1-Dimensional CNN-Bidirectional-LSTMs with Residuals shortly known as Res-CNN-BiLSTM. In this paper we have proposed the architecture and compared its performance with different approaches of Embedding Deep Learning Algorithms.
\end{abstract}


\section{Introduction}
The disturbance in the mental health has many facets and one factor cannot determine it. Since social media access has became easily available to masses, it has became a cause for mental health disturbance at an accelerated rate. Exploitation is done on regular basis by many social media users which actually leads to cyberbullying \cite{Nixon2014CurrentPT}. The cyberbullying can lead to severe effects in person's lifestyle, especially adolescents. The cyberbullying can be based on religion, age, gender and ethnicity. Expression of opinions on social media definitely leads to conflict when sentiments of the reader are offended. As a matter of fact, the reasons are very varied for offense and the readers can take it personally on many different levels, cyberbullying is done as a retaliation. This sometimes can cause a mental disturbance if the person is genuinely looking for suggestions or validation on opinions. The perspectives given on the race even if pointing out the obvious can cause cyberbullying with racism. The religion and ethnicity are divisions of racism. Religion turns out to be a major factor as it is directly based on the principles followed by a certain community. These principles not necessarily can be acceptable for the ardent followers of their religion which eventually can turn out to be a cause for cyberbullying. The basis of any religion is based on the belief system of principles and definitely posting an opinion about it on social media is acceptable, but since sentiments are attached to religion, it eventually can end up worse. Same is the case with ethnicity. It is associated on nation that works on similar beliefs. Not everyone accepts the belief system of other ethnicity and wants to place an opinion on the same. Religion cannot be termed on a physical level, but ethnicity does. The physical level cyberbullying commonly done is on the age and gender factors. The age is the factor where people are reluctant towards the fact that things that are not achievable based on the age are being achieved by the people that do not follow the social norms. The main psychological reason that promotes cyberbullying is done under social conventions. The social conventions are sub-consciously planted in the minds of people at very early age. Certain factors are not accepted by the people who do not follow similar social convention. Humans psychologically work with anecdotes and acceptable behaviour that considers the factors of race, religion, color, ethnicity and gender. The basis for argument of bullying is clearing when the two people have disparity in the thought process and since social media has became an easy platform for expression of opinions against some norms, it results into conflicts. Gender plays a major role in cyberbullying \cite{Rego2018ChangingFA}. Misogyny is the main reason for it. People are fixated on the roles played by gender and when the roles are reversed, the advent expression of opinion steps in every person with the granted power of social media. The interference in the conflicts is essential because the results can eventually get worse. This probably can get solved when it is a physical confrontation but really a difficult problem when handling online. This constant abuse does effect mental health and a detection of it in early stages proves to be an efficient problem. The biggest instrument of liberation is Twitter for expression of opinions. It is often used for cyberbullying and harassment \cite{matias2015reporting}. Twitter no doubt gave freedom of speech on next level for users but definitely is used as a tool for misuse by many people. Psychologically people are inclined towards expressing their opinions on platform that does not consider any restrictions of gender, race, ethnicity and age. Social media for sure leveraged such freedom yet created an issue too. This is an intricate topic and can be understood from the Ethical Neutrality Principle \cite{10.2307/25071761,10.2307/3710463}. This can be better understood from the example that knife was a powerful tool that changed the civilization, but it depends how one uses it. It can be used for daily chores as well as can be used to hurt someone. Same is the case with Twitter. The power of twitter has obviously helped many people express opinions without hesitation and many have fell victim in the trap of cyberbullying. Such form of sensitive issue needs to be addressed with care because it can help in preservation of traumas that promote mental health disturbance.

The problem of mental health disturbance can be targeted by directly managing the cyberbullying. This definitely is a Natural Language Processing \cite{collobert2011natural} problem as text is the main medium of communication between 2 people. Analyzing the text and then deriving the semantics \cite{wang2021semantic} from the text can be done. This semantics can be derived using Deep Learning \cite{lecun2015deep} process. The applications of Deep Learning have grown in the area of NLP at a very great lengths \cite{torfi2020natural}. The approach that we used in this paper will be explained effectively in consecutive sections in detail. We managed to tackle the issue of deriving type of cyberbullying in the tweet using the Deep Learning with NLP approach.

\section{Preliminary Approach}
This section of the paper will give you a notion on tackling the problem of Natural Language Processing using Deep Learning. The elements required to understand the approach we have represented, so consider this as a review of the methods.

\subsection{Word Embedding}
The main problem of the text data is to derive the semantics of the data for which many methods have been formulated. The text holds information for which the algorithm needs to understand the language. Some primordial methods like Bag of Words \cite{Zhang2010UnderstandingBM} and TF-IDF \cite{ref1} were used but have limitations which were implemented and proven \cite{tambe2022effects} in detail. These represent the data in a matrix form yet are limited by the region space. The concept was Word Embedding \cite{almeida2019word} was introduced later which has a very high dimensional space and trained with language modeling \cite{jing2019survey}. The process includes a dataset to processed with one hot encoding \cite{Hancock2020SurveyOC}. After that a vector is considered with very high dimensional space. These have a certain seed value initially. The initialized vectors are later on mapped with all the one hot encoded values. The training is very extensive and computationally expensive. The dimensional space of such a word embedding cannot be visualized in the 2-dimensional space. It ranges over 100 dimensions. Many models have around 300 dimensions, which is the only reason for extensive training for longer periods.

\subsection{Word2Vec}
In order to improve the current state of the word embeddings, a powerful statistical model developed, name Word2Vec \cite{mikolov2013efficient}. The representation of the words was done on the next level in word2vec. The model was able to derive subtle differences very efficiently. Representations directly effect the semantics as the relation of the word states understanding of the language. 2 different types of models were used for word2vec known as Continuous Bag-of-Words and Continuous Skip-Gram Model. The continuous bag-of-words model derives the embedding by prognosticating the current word based on its context. It given by formula
\begin{equation}
J_\theta = \frac{1}{T}\sum\log{p}\left(w_{t}\mid{w}_{t-n}\ldots,w_{t+n}\right)
\end{equation}

The continuous skip-gram model derives by prognosticating the surrounding words given in current word. This model is efficient in training as compared to the word embedding model. It is given by formula
\begin{equation}
    J_\theta = \frac{1}{T}\sum\sum\log{p}\left(w_{j+1}\mid{w_{t}}\right)
\end{equation}

\subsection{GloVe}
Many statistical models were developed for the concept of Topic Modeling \cite{Mahmood2013LiteratureSO}. This has a prominent algorithm known as Latent Semantic Analysis \cite{Reidy2009AnIT} that use matrix factorization. This can help derive very good semantics but had some limitations when compared with word2vec. Later in order to develop an efficient model that combines best of both worlds, Global Vectors for Word Representation also known as GloVe \cite{pennington2014glove} was developed. This model is generally better than the word2vec in most of the cases. It works with entire corpus using word context matrix and word co-occurrence matrix.

\subsection{Recurrent Neural Network}
The important aspect of the natural language processing problem is that it is sequential data \cite{Dietterich2002MachineLF}. For learning representations from such type of data in deep learning, one has to use sequence models. The type of neural network used for sequential data is Recurrent Neural Network \cite{sherstinsky2020fundamentals}. The data is given in  a sequence of $x_t$. This sequence is given to activation layer that is represented by $a_t$. This can be represented in an equation by
\begin{equation}
    a_t = (W.h_{t-1}) + (U.x_t) + b
\end{equation}
where $W$ are the initialized weights from the previous layer, $h_{t-1}$ are hidden units from the layer, $U$ are the initialized weights for the current layer, $x_t$ is the current sequential data and $b$ is bias. This activation layer is passed through \textit{tanh} \cite{227257} function. This helps to learn representations for the network and evaluates the values of activation layer in the range of [-1, 1]. This process can be represented by the $h_t$ and is given to output layer. The equation of $h_t$ can be represented as $h_t = tanh(a_t)$. This output layer is represented by $o_t$. This can be given by equation
\begin{equation}
    o_t = (V.h_t)+c
\end{equation}
where $V$ are the weights initialized and $c$ is the bias for the layer. The computation of gradients done is with respect to every initialization. This process is known as back-propagation \cite{Rumelhart1986LearningRB} and is used to regain lost information from the forward pass of the network.

\subsection{Long Short Term Memory Network}
The Recurrent Neural Network was definitely successful but had some drawback. While training a very long sequence of data, the gradient calculation becomes a very difficult task. The gradients become very negligible for computation and this problem is known vanishing gradient problem \cite{10.1142/S0218488598000094}. In order to tackle this problem Long Short Term Memory Network abbreviated as LSTM \cite{Hochreiter1997LongSM, olah2015understanding, staudemeyer2019understanding} was developed. It has potential to handle a very long sequence of data. The LSTM is designed using varied gates. The main thing responsible for LSTM is the cell state. This holds the information for particular cell. The various gates are forget gate, input gate, output gate. The forget gate is very straight forward and is used for filtering out the important information. It decides which information needs to be kept and which is supposed to be forgotten. It uses a sigmoid \cite{Cybenko1989ApproximationBS} function that is represented between [0,1]. The values are very straight forward where 0 indicates forget every single thing and 1 indicates keep all the information. The equation of the forget gate can be given by
\begin{equation}
    f_t = \sigma(W_f.[h_{t-1}, x_t] + b_f)
\end{equation}
where $f_t$ indicates forget gate for state $t$, $W_f$ states weights initialized for forget gate, $h_{t-1}$ states output of previous state, $x_t$ states input from current state, $b_f$ indicates weights initialized for bias of forget gate and $\sigma$ states the sigmoid function. Maintaining the information of the forget gate, we have to consider the input gate. The input gate has 2 aspects, the input state and the candidate values. The input state is very similar to the forget gate equation and uses the sigmoid function. The equation is given by
\begin{equation}
    i_t = \sigma(W_i.[h_{t-1}, x_t] + b_i)
\end{equation}
where $i_t$ indicates input gate for state $t$, $W_i$ states weights initialized for input gate, $h_{t-1}$ states output of previous state, $x_t$ states input from current state, $b_i$ indicates weights initialized for bias of input gate and $\sigma$ states the sigmoid function. The candidate value on the other hand is a vector of information for the current state that helps learn LSTM representations. The candidate value uses tanh function. The equation is given by
\begin{equation}
    c_t = tanh(W_c.[h_{t-1}, x_t] + b_c)
\end{equation}
where $c_t$ indicates candidate value for state $t$, $W_c$ states weights initialized for candidate value, $h_{t-1}$ states output of previous state, $x_t$ states input from current state, $b_c$ indicates weights initialized for bias of candidate value and this is enclosed within tanh activation function. Now calculation of the representation is done by multiplying input gate and candidate value. This is the information of the current input state. But it is supposed to be calculated with the consideration of the forget gate that was calculated earlier. The forget gate is multiplied with the previous cell state and current input state is added to it. This will gives the current cell state for which the equation is given by
\begin{equation}
    C_t = f_t*C_{t-1} + i_t*c_t
\end{equation}
where $C_t$ is the current cell state, $f_t$ is the forget gate which is multiplied with $C_{t-1}$ which is previous cell state and added with multiplication of $i_t$ input gate and $c_t$ candidate value. Now we consider the output gate which gives the final output represented by $h_t$. This output gate is first used for calculation of the data from the states, then is multiplied with current cell state that is enclosed within $tanh$ activation function. This gives the best possible information to be preserved and given as output. The equation is given by
\begin{displaymath}
  o_t = \sigma(W_o.[h_{t-1}, x_t] + b_o)
\end{displaymath}
and final output is given by
\begin{equation}
    h_t = o_t * tanh(C_t)
\end{equation}
where $o_t$ indicates output gate for state $t$, $W_o$ states weights initialized for output gate, $h_{t-1}$ states output of previous state, $x_t$ states input from current state, $b_o$ indicates weights initialized for bias of output gate and this is enclosed within $sigmoid$ activation function. The final is output represented by $h_t$ and is achieved using multiplication of output gate $o_t$ and current cell state $C_t$ enclosed within a $tanh$ activation function.

\subsection{Bidirectional Networks}
The concept of probabilistic models \cite{10.5555/944919.944966} was a very important observation for many researchers. The Hidden Markov Model \cite{1165342} is a powerful probabilistic model. Inculcation of dynamic programming \cite{1098755} principles in it makes it very efficient. The look ahead ability of the hidden Markov models was required for RNN. Since the data is sequential, the RNNs can implement the same principle where they look ahead of the sequence and get more information out of the model. The Bidirectional Recurrent Neural Network \cite{650093} was developed for such reason. The concept of it is that it adds hidden layer along with original layers that retains information from the future sequence. This enforces the principles of dynamic programming of hidden Markov models. This similar concept was further extended for the LSTMs known as Bidirectional Long Short Term Memory Network abbreviated as BiLSTMs \cite{huang2015bidirectional}. This gives very good precision in many cases over the traditional state-of-the-art LSTM models.

\section{Methodology}
This section focuses on the proposed architecture for this paper. It uses all the concepts described in the earlier section. Obviously some extra features are being added to the architecture which will be explained in this section.

\subsection{Word Embedding}
\begin{figure}[hbt]
  \centering
  \includegraphics[scale=0.5]{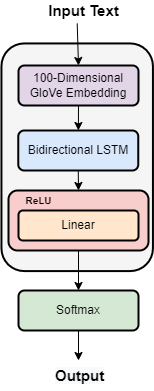}
  \caption{GloVe Model with Bidirectional LSTM}
  \label{fig:a}
\end{figure}

The model architecture is divided into various parts and this is one of the parts. The Embeddings that we used in our proposed architecture are 100-Dimensional GloVe Embeddings. These are trained over 6 billion Tokens and 400k vocab. The parameters learned from this layer are 2706400. Followed by this layer is a Bidirection LSTM layer. The units used 512 for LSTM and parameters learned are 2510848. Followed by this is a Linear layer with 32 hidden units and applied with ReLU \cite{agarap2019deep} activation function in which parameters learned are 32800. The final output layer is then achieved with 5 hidden units and softmax \cite{liu2017largemargin} activation function. The total parameters learned from the network are over 5.2 million.

\subsection{Character Level Embedding}
The word embedding has some intricate limitations. No matter how better model one trains, sometime or the other user will get a word that is not available in the vocabulary. In such a case the word will termed out of vocabulary word. Developed algorithm like GloVe will term it as out of vocabulary and assign some random vectorized value. 

\begin{figure}[hbt]
  \centering
  \includegraphics[angle=90, scale=0.5]{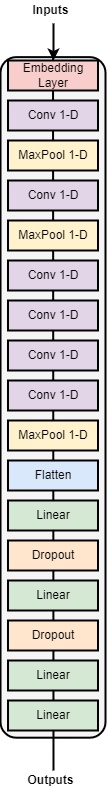}
  \caption{1-Dimensional CNN}
  \label{fig:b}
\end{figure}

This issue was observed and character level embedding \cite{zhang2016characterlevel} was taken into consideration. It uses 1-dimensional convolutional neural network \cite{KIRANYAZ2021107398} for representations at character levels. The importance of character level embedding can be given by an example. Consider word like \textit{unconfident} which has a literal meaning in English language. But now consider the word \textit{inconfident}. The word literally does not exist, but the \textit{in} prefix before the word confident clearly gives you a notion of word representation. Character level embedding gives you this effect by extracting the shorter segments from long sequences of data. Our proposed model has one embedding layer for character embeds that takes input sequence. First 1-D convolutional layer is applied with 256 filters of $7x7$ size with $3x3$ strides with ReLU activation function which learns 123904 parameters. Then a 1-D Max Pooling Layer is applied with $3x3$ dimension. Again similar type of 1-D convolutional layer is applied that learns 459008 parameters. Then one more 1-D Max Pooling Layer is applied for that results in reduction. Then 4, 1-D convolutional layers are applied with same constraints. These learn 196864, 196864, 196864, 196864 parameters respectively. Then one 1-D Max Pooling Layer is applied for reduction. Then all the process is Flattened out in a fully connected layer. Then one Linear layer is applied with 1024 hidden neurons and ReLU activation function which learns 8913920 parameters. Then a Dropout layer is applied with 0.5 threshold. Then this process of Linear layer and Dropout layer is once again repeated where linear layer learns 1049600 parameters. Then we apply one linear layer with 32 hidden neurons and ReLU activation that learns 32800 parameters. Finally we apply the last linear layer with 5 hidden neurons that uses Softmax activation function and we get the output probabilities. The total parameters learned by network are over 11.3 million.

\subsection{Combining Both Networks}
\begin{figure}[hbt]
  \centering
  \includegraphics[scale=0.8]{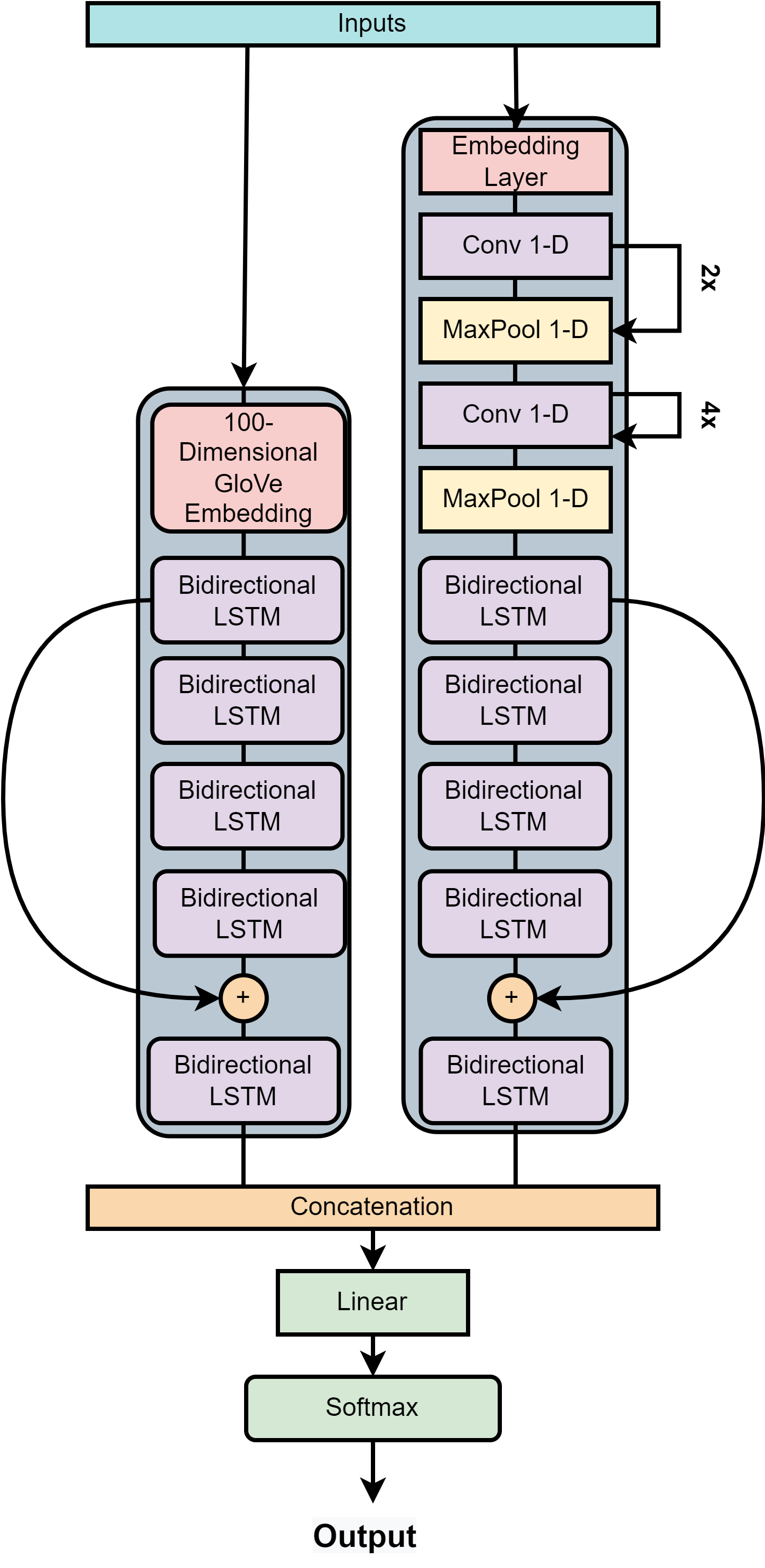}
  \caption{Combined Network}
  \label{fig:c}
\end{figure}

So far now we were building a very strong basic conceptualization of the topics required to understand our proposed architecture. The architecture we proposed is given in Figure \ref{fig:c}. The word embeddings model and character embeddings model are combined to get the best of both worlds. Since the networks are explained in depth in above sections, the emphasis should be given to changes. We believe the combination of both the networks yield a good result covering the leaks of each other. The architectural changes indicate removing the output layer of both the networks for purpose of combining them. The Bidirectional LSTM with 64 hidden units are generated for both the networks. Then these layers are concatenated. The process of concatenation does yield better result in many cases. The concatenation does require the same dimension of both layers. The concatenation is then given to a linear layer with 32 hidden neurons with ReLU activation function. This then connects to final linear layer with 5 hidden neurons and softmax activation function for output. The total number of parameters learned by network are over 48.8 million. The idea is to train 2 independent networks simultaneously and then combine them to yield a different more efficient network. Similar type can be network seen for solving the Named Entity Recognition problem \cite{chiu2016named}. We managed to make this work for Classification problem for mental health preservation. In our architecture we have also utilized the concept of Skip Connections \cite{wu2020skip} from Residual Networks \cite{he2015deep}. The 4 Bidirectional LSTMs involved in this residual process have 512 hidden units. This is done for deriving the best semantics out of learnt embedding representations. The understanding of semantics is the process done by the neural networks.

\subsection{Data}
The data we needed was a very specific featured based. We wanted  dataset that can give us the insights of cyberbullying on various different levels. For identification of the type of bullying, we wanted to specifically target the major divisions in the bullying. Religion, Age, Gender and Ethnicity are the major divisions we targeted. We were able to gain data of 47000 tweets that were related to cyberbullying \cite{9378065}. This data had major labels we wanted to target. Although a lot of preprocessing was required for sure. The dataset had labels equally distributed, but had duplicate records which we tackled in feature engineering \cite{7506650}. The text wanted preprocessing on an intricate level, where since tweets were involved, emoticons were involved. We also preprocessed stop-words \cite{sarica2021stopwords} from corpus. Label Encoding the labels was also necessary as the labels were categorical variables. 

\section{Results}
This section focuses on the performance of our network with respect to Character Level Embeddings Model and Word Representations Model.

\subsection{Analysis}
The first step during the phase of preprocessing is analysis of data. This mainly emphasizes on the visualization aspect. The word representations are very high dimensional and beyond the scope of visualization. For a similar reason we used some techniques. The only technique for visualizing a very high dimensional embedding in high dimensional spaces is with dimensionality reduction \cite{Maaten2009DimensionalityRA} algorithms. 

\begin{figure}[hbt]
  \centering
  \includegraphics[scale=0.25]{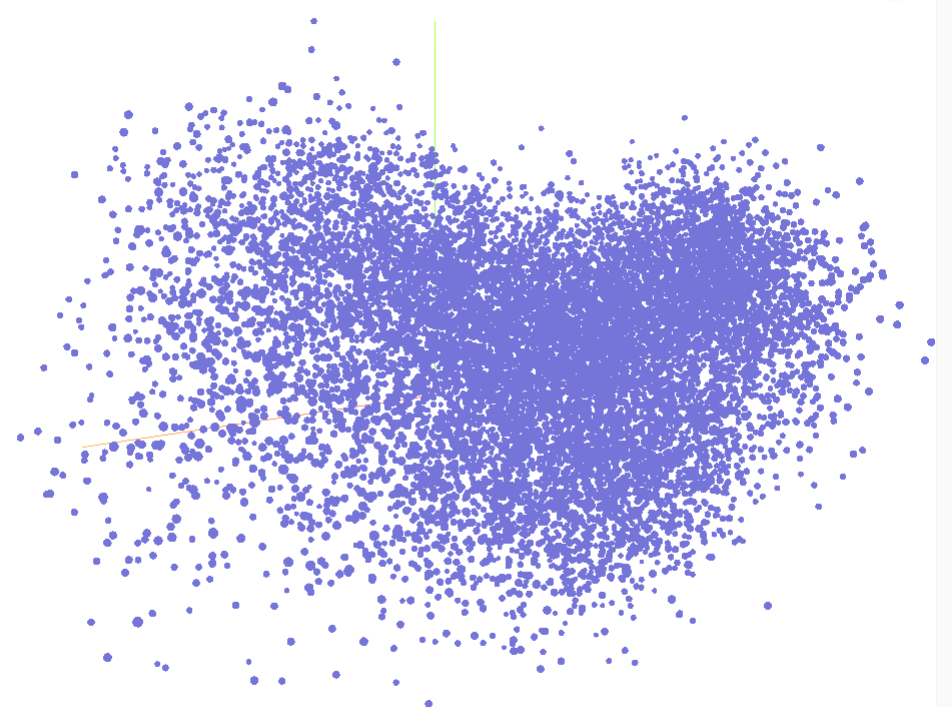}
  \caption{Word Embedding Projection using PCA}
  \label{fig:g}
\end{figure}
\begin{figure}[hbt]
  \centering
  \begin{tabular}{ll}
  \includegraphics[scale=0.15]{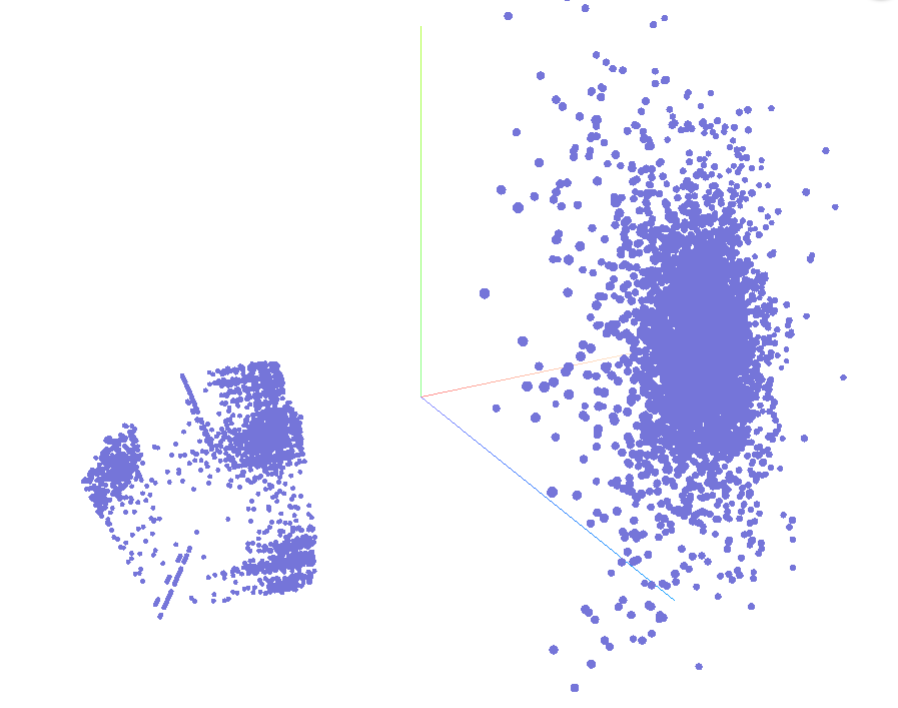}
  &
  \includegraphics[scale=0.15]{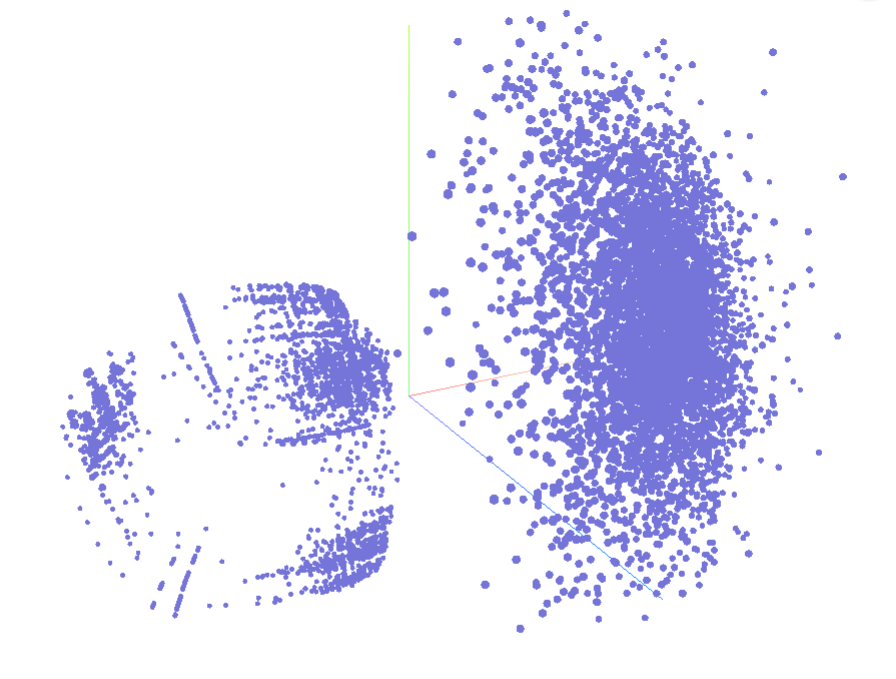}
  \end{tabular}
  \begin{tabular}{ll}
  \includegraphics[scale=0.15]{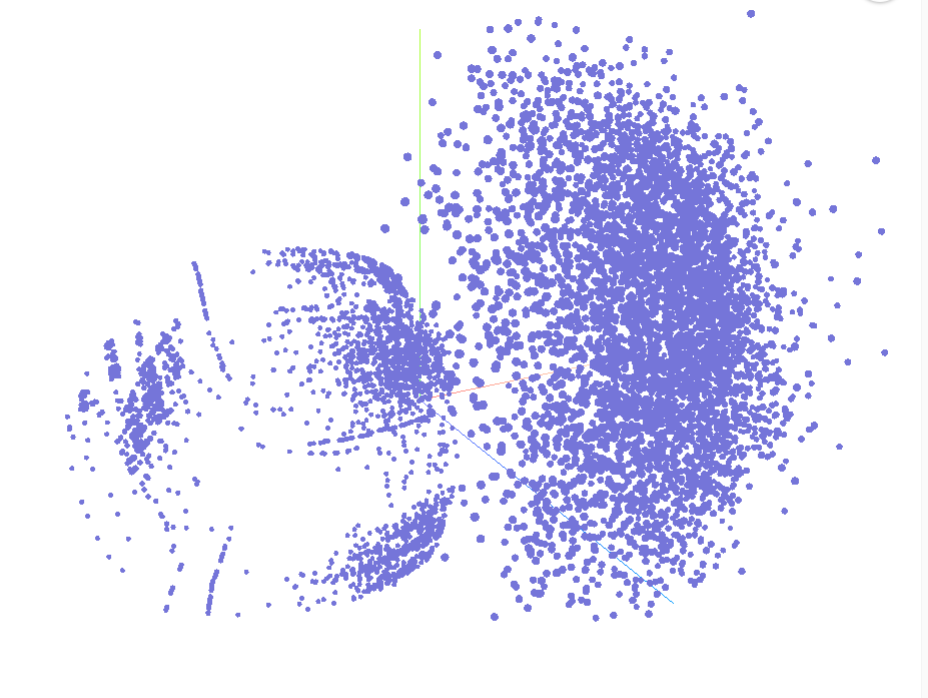}
  &
  \includegraphics[scale=0.15]{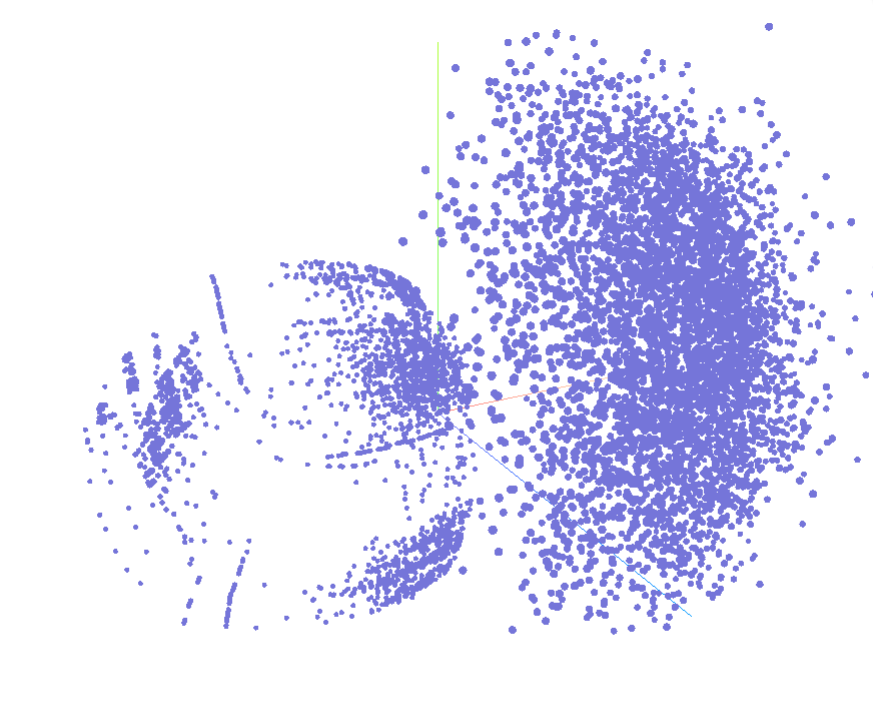}
  \end{tabular}
  \caption{Phases of the t-SNE for Word Representation}
  \label{fig:h}
\end{figure}

The embedding can be projected in lower dimension using Principal Component Analysis \cite{Jolliffe2016PrincipalCA}. It is an unsupervised dimensionality reduction algorithm. The algorithm uses Singular Value Decomposition \cite{1102314}. The Figure \ref{fig:g} gives a detailed PCA visual representation of the GloVe word embedding. One another approach for word embedding representation is by using t-SNE \cite{JMLR:v9:vandermaaten08a} Algorithm. This is also a type of dimensionality reduction method designed for visualizing very high dimensional data in visualization plane. The visualization for word embedding using t-SNE can be done in phases. The Figure \ref{fig:h} gives the t-SNE visualization for GloVe Word Embedding.

\subsection{Precision}
The Precision is a metric that gives the amount of positive prediction that were correct when compared with the labels. The Precision requires True Positive and False Positive.

\begin{table}[hbt]
  \centering
  \caption{Macro and Weighted Precision}
  \label{tab:a}
  \begin{tabular}{ccl}
    \toprule
    Algorithm&Macro Precision&Weighted Precision\\
    \midrule
    1-D CNN & 91\% & 91\%\\
    GloVe & 92\% & 92\%\\
    Res CNN-BiLSTM & 92\% & 92\%\\
  \bottomrule
\end{tabular}
\end{table}

The Table \ref{tab:b} gives insights of precision for each algorithm with respect to every single label. In such a scenario the scores in some cases is ahead of our proposed model, but important thing to note is the variation is less. That shows the class level precision is very good for our system.

\begin{table}[hbt]
    \centering
  \caption{Class Level Precision}
  \label{tab:b}
  \begin{tabular}{cccl}
    \toprule
    Classes & 1-D CNN & GloVe & ResCNN-BiLSTM\\
    \midrule
    Class 0 & 97\% & 95\% & 94\%\\
    Class 1 & 95\% & 97\% & 96\%\\
    Class 2 & 96\% & 99\% & 98\%\\
    Class 3 & 87\% & 89\% & 91\%\\
    Class 4 & 78\% & 77\% & 81\%\\
  \bottomrule
\end{tabular}
\end{table}

\subsection{Recall}
Recall is a metric that calculates the actual positives from test class in a correct fashion. Just like Precision it also uses the pillars of confusion matrix. Just it equates with false negatives. This also can be represented on a macro and weighted recall for overall labels. 

\begin{table}[hbt]
    \centering
  \caption{Macro and Weighted Recall}
  \label{tab:c}
  \begin{tabular}{ccl}
    \toprule
    Algorithm&Macro Recall&Weighted Recall\\
    \midrule
    1-D CNN & 91\% & 91\%\\
    GloVe & 92\% & 92\%\\
    Res CNN-BiLSTM & 92\% & 92\%\\
  \bottomrule
\end{tabular}
\end{table}

The recall values from Table \ref{tab:c} are similar to the precision values. But just like precision, recall also will be considered for class level predictions. Inference derived from Table \ref{tab:d} clearly states a better result of GloVe Embeddings as compared to class level precision. 1-D CNN obviously has higher variation in all the classes. GloVe has less variation but still our proposed system is able to give better recall predictions.

\begin{table}[hbt]
    \centering
  \caption{Class Level Recall}
  \label{tab:d}
  \begin{tabular}{cccl}
    \toprule
    Classes & 1-D CNN & GloVe & ResCNN-BiLSTM\\
    \midrule
    Class 0 & 92\% & 94\% & 94\%\\
    Class 1 & 98\% & 98\% & 98\%\\
    Class 2 & 98\% & 97\% & 98\%\\
    Class 3 & 89\% & 86\% & 88\%\\
    Class 4 & 76\% & 84\% & 81\%\\
  \bottomrule
\end{tabular}
\end{table}

\subsection{$F_1$-Score}

The $F_1$-Score is a metric designed using Precision and Recall. It gives the accuracy value of the model on the dataset. The metric for this can be represented with values calculated from all the different networks used in this paper. The formula for $F_1$-Score is given by

\begin{equation}
    F_1 = 2*\frac{Precision*Recall}{Precision+Recall}
\end{equation}

The Table \ref{tab:e} gives the $F_1$-Score on different models from this paper. Just like Precision and Recall, the $F_1$-Score can also be calculated on class level labels.

\begin{table}[hbt!]
    \centering
  \caption{Macro and Weighted $F_1$-Score}
  \label{tab:e}
  \begin{tabular}{lll}
    \toprule
    Algorithm&Macro $F_1$&Weighted $F_1$\\
    \midrule
    1-D CNN & 91\% & 91\%\\
    GloVe & 92\% & 92\%\\
    Res CNN-BiLSTM & 92\% & 92\%\\
  \bottomrule
    \end{tabular}
\end{table}

The metric of $F_1$-Score gives insights on scores of different networks used. The variation can been seen in 1-D CNN Character Level Embedding. GloVe Word Embeddings Model and our proposed model are very close to each other. The Class 0 and Class 1 for GloVe gives a better score and our model gives better result for Class 3 and Class 4.

\begin{table}[hbt!]
  \caption{Class Level $F_1$-Score}
  \centering
  \label{tab:f}
  \begin{tabular}{cccl}
    \toprule
    Classes & 1-D CNN & GloVe & ResCNN-BiLSTM\\
    \midrule
    Class 0 & 94\% & 95\% & 94\%\\
    Class 1 & 97\% & 98\% & 97\%\\
    Class 2 & 97\% & 98\% & 98\%\\
    Class 3 & 88\% & 88\% & 89\%\\
    Class 4 & 77\% & 80\% & 81\%\\
  \bottomrule
\end{tabular}
\end{table}

\section{Conclusion}

Mental Health Preservation is the main aim of this paper, where we specifically targeted cyberbullying that causes exploitation using social media as a medium. Cyberbullying can be done in a various manners, related to Religion, Ethnicity, Age and Gender. In this paper we specifically emphasized on such sensitive issue and leveraged the Natural Language Processing domain with Deep Learning. Natural Language Processing domain is used because the data collected is in the text format. Along with NLP to derive semantics on a good level, we used Deep Learning. In this paper we proposed Hybrid Deep Learning Model of 1-Dimensional Convolutional Neural Networks and Bidirectional LSTM with Residuals. We wrote this paper in such a manner that it also emphasizes on the prerequisite aspects so that naive researchers can also get a walk-through of the process to reach to the state of our model. We consider this work as a review with proposed system that aims all the NLP researchers. Obviously many more thought processes will be promoted further from our work and we would be glad to be a helping hand in journey of young researchers through this paper.

\section*{Acknowledgments}
We would like to thank Andrew Maranhão for providing the dataset on the Kaggle Platform named Cyberbullying Classification. We have cited the dataset as prescribed by the author.

\bibliographystyle{unsrt}  
\bibliography{references}

\end{document}